\documentclass[english,runningheads,a4paper]{llncs}[2018/03/10]

\usepackage{latexsym}
\usepackage{url}
\usepackage{multirow}
\usepackage{amssymb}
\usepackage{graphicx}
\usepackage{amsmath}
\usepackage{floatrow}

\usepackage{regexpatch}
\makeatletter
\edef\switcht@albion{%
  \relax\unexpanded\expandafter{\switcht@albion}%
}
\xpatchcmd*{\switcht@albion}{ \def}{\def}{}{}
\xpatchcmd{\switcht@albion}{\relax}{}{}{}
\edef\switcht@deutsch{%
  \relax\unexpanded\expandafter{\switcht@deutsch}%
}
\xpatchcmd*{\switcht@deutsch}{ \def}{\def}{}{}
\xpatchcmd{\switcht@deutsch}{\relax}{}{}{}
\edef\switcht@francais{%
  \relax\unexpanded\expandafter{\switcht@francais}%
}
\xpatchcmd*{\switcht@francais}{ \def}{\def}{}{}
\xpatchcmd{\switcht@francais}{\relax}{}{}{}
\makeatother

\usepackage{ifluatex}
\ifluatex
  \usepackage{fontspec}
  \usepackage[english]{selnolig}
\fi

\iftrue 
  \ifluatex
  \else
    \usepackage[%
      rm={oldstyle=false,proportional=true},%
      sf={oldstyle=false,proportional=true},%
      tt={oldstyle=false,proportional=true,variable=false},%
      qt=false%
    ]{cfr-lm}
  \fi
\else
  \ifluatex
  \else
    \usepackage{newtxtext}
    \usepackage{newtxmath}
    \usepackage[zerostyle=b,scaled=.9]{newtxtt}
  \fi
\fi

\ifluatex
\else
  \usepackage[T1]{fontenc}
  \usepackage[utf8]{inputenc} 
\fi

\usepackage{graphicx}
\usepackage{upquote}

\usepackage{booktabs}

\usepackage{paralist}

\usepackage{csquotes}

\usepackage{textcmds}
\usepackage{dirtytalk}
\RequirePackage[%
  babel,%
  final,%
  expansion=alltext,%
  protrusion=alltext-nott]{microtype}%
\DisableLigatures{encoding = T1, family = tt* }

\usepackage{url}
\makeatletter
\g@addto@macro{\UrlBreaks}{\UrlOrds}
\makeatother

\usepackage{xcolor}

\usepackage{listings}
\usepackage{chngcntr}
\AtBeginDocument{\counterwithout{lstlisting}{section}}

\usepackage[%
  square,        
  comma,         
  numbers,       
  sort&compress, 
]{natbib}

\ifluatex
\else
  \SetExpansion
  [ context = sloppy,
    stretch = 30,
    shrink = 60,
    step = 5 ]
  { encoding = {OT1,T1,TS1} }
  { }
\fi

\usepackage{stfloats}
\fnbelowfloat

\usepackage{hyperref}
\hypersetup{hidelinks,
  colorlinks=true,
  allcolors=black,
  pdfstartview=Fit,
  breaklinks=true}
\usepackage[all]{hypcap}

\ifluatex
\else
  \input glyphtounicode
\fi

\usepackage[tableposition=top]{caption}
\usepackage{floatrow}
\begin{document}
\title{Robust Evaluation of Language--Brain Encoding Experiments}
\author{Lisa Beinborn, Samira Abnar, Rochelle Choenni}
\authorrunning{L. Beinborn et al.}
\institute{Institute for Logic, Language and Computation\\
  University of Amsterdam\\
  \email{l.beinborn@uva.nl}, \email{s.abnar@uva.nl}, \email{rochelle.choenni@student.uva.nl}}
\maketitle
\begin{abstract}
Language-brain encoding experiments evaluate the ability of language models to predict brain responses elicited by language stimuli. The evaluation scenarios for this task have not yet been standardized which makes it difficult to compare and interpret results. We perform a series of evaluation experiments with a consistent encoding setup and compute the results for multiple fMRI datasets. In addition, we test the sensitivity of the evaluation measures to randomized data and analyze the effect of voxel selection methods. Our experimental framework is publicly available to make modelling decisions more transparent and support reproducibility for future comparisons.
\end{abstract}

\section{Introduction}
Representing language in a computationally usable format has been a research goal since the beginning of computational linguistics. In the last decade, distributional representations which interpret words, phrases, sentences, and even full stories as a high-dimensional vector in semantic space have become the most common standard. These representations are obtained by training language models on large corpora to optimally encode contextual information.

The quality of language representations is commonly evaluated on a set of downstream tasks. These tasks are either driven by engineering adequacy (e.g. the effect of the language representations on the performance of systems such as machine translation) or by the ability to reproduce human decisions (e.g. the performance of the representations on semantic similarity or entailment tasks). Many language researchers, however, are driven by the urge to better understand the underlying principles of human language processing. 

With the increasing availability of brain imaging data, it has become popular to evaluate computational models by their ability to simulate brain signals related to human language processing \cite{Hale2018,Jain2018,Pereira2018}. If we can develop models that encode linguistic information in a way that is comparable to the activity in human brains, we will get one step closer to cognitively plausible models of human language understanding. While experimenting with human brains is evidently strictly constrained and regulated due to ethical reasons, we can easily query, adapt, constrain, degrade, and manipulate the computational model and analyze the effect on its language processing capabilities.

Although working with brain imaging data is highly promising from a cognitive perspective, it comes with many practical limitations. Brain datasets are usually too small for powerful machine learning models, the imaging technology produces noisy output that needs to be adjusted by statistical correction methods, and most importantly, only very few datasets are publicly available. Experiments in previous work are usually performed on a single dataset, so that it is unclear whether the observed effects are generalizable. In addition, the applied evaluation procedures have not yet been standardized. Understanding the subtle differences in the experimental setup to interpret the results can be particularly difficult because it has not yet become a common practice to publish the experimental code along with the results.

To the best of our knowledge, this paper provides the first analysis of language--brain encoding experiments which applies a consistent evaluation scenario across multiple fMRI datasets. We examine whether different evaluation measures provide different interpretations of the predictive power of the encoding model. Our experimental framework is publicly available to make modelling decisions more transparent and facilitate reproducibility for future comparisons. Due to its modular architecture, the pipeline can easily be extended to experiment with other datasets and language models.\footnote{The code is available at \url{https://github.com/beinborn/brain-lang}}

{\setlength{\tabcolsep}{2pt}
\begin{table}[ht]
\centering
\small
\caption{4 fMRI datasets for language--brain encoding. In \textsc{Words} and \textsc{Stories}, stimuli have been isolated by averaging over the brain responses. The \textsc{Alice} and \textsc{Harry} datasets contain continuous stimuli.}
\label{tbl_braindata}
\begin{tabular}{lllrrrr} 
\toprule
 Name &Stimuli& Presentation mode& Subj. & Scans & Voxel size & Reference \\
\midrule
 \textsc{Words}  & 60 words & Word + image & 9 &360 & 3x3x6& \citet{Mitchell2008}\\
\textsc{Stories}  & 40 stories & Read sentences & 30 & 40 & 3x3x3& \citet{Dehghani2017}\\
\midrule
\textsc{Alice} & 1 chapter   & Listen to audio book  &27 &362& 3x3x3& \citet{Brennan2016}\\
 \textsc{Harry} & 1 chapter & Read word by word & 8& 1351 & 3x3x3 & \citet{Wehbe2014}\\
\bottomrule
\end{tabular}
\end{table}
}
\section{Human-centered Evaluation of Computational Models}
As computational language models are trained on human-generated text, their performance is inherently optimized to simulate human behavior. Although novel architectural solutions attract notable interest in the research community, the ultimate benchmark for a model is the ability to approximate human language processing abilities. Models are supposed to reach a gold standard of human annotation decisions \cite{Resnik2010} and the difficulty of a task is often estimated by the inter-annotator agreement \cite{Artstein2008} or by error rates of human participants \cite{Beinborn2014}. While these product-oriented evaluations focus on a final outcome, procedural measures of response times \cite{Monsalve2012} or eye movements \cite{Barrett2018} are analyzed to provide deeper insights on sequential phenomena like attention or processing complexity. As neural network models are inspired by neuronal activities in the human brain, it is particularly interesting to analyze similarities and differences between distributed computational representations and low-level brain responses. 

Electroencephalography (EEG) measures can be used to study specific semantic or syntactic phenomena \cite{Hale2018,Fyshe2016,Sudre2012} and compare the processing complexity of computational models to brain responses, for example, with respect to the N400 and P600 effects \cite{Frank2013}. Signals with higher spatial resolution like magnetoencephalography (MEG) and functional magnetic resonance imaging (fMRI) are often used for experiments which are known as brain decoding and brain encoding. In the decoding setup, a computational model learns to identify differences in the signal and to discriminate between the responses for abstract and concrete words \cite{Anderson2017}, for different syntactic classes \cite{Bingel2016,Li2018}, for levels of syntactic complexity \cite{Brennan2016}, and many other linguistic categories. \citet{Mitchell2008} have shown that it is not only possible to distinguish between semantic categories, but that a model can even learn to distinguish which word a participant is reading. The reverse direction of predicting the brain response that would most likely be observed for a novel linguistic stimulus is commonly called encoding. The encoding task requires a strong computational representation of the stimulus that reflects the shared properties of different stimuli and the relations between stimuli. For the remainder of this paper, we will focus on the language--brain encoding task and on fMRI datasets. 

Many word representations have been tested on the \citet{Mitchell2008} data including information from lexical resources, distributional, and multimodal representations \cite{Abnar2017,Anderson2017,Bulat2017,Xu2016}. It has also been proposed to directly feed the brain signal into the language model as an additional source of information \cite{Athanasiou2018,Fyshe2014}. Recently, new approaches for encoding and decoding of datasets using longer linguistic stimuli such as sentences \cite{Pereira2018} and even full stories \cite{Jain2018,Dehghani2017,Brennan2016,Wehbe2014} are emerging. In some experiments, it has been shown that contextualized representations obtained from recurrent neural networks \cite{Wehbe2014align,Jain2018} seem to represent the continuous stimuli slightly better than models that represent sentences as a conglomerate of context-independent word representations \cite{Dehghani2017,Pereira2018}. However, these results are hard to generalize because they have been tested only on a single dataset. \citet{Gauthier2018} raise doubts about the informativeness of encoding results because differences between models are not reflected. Our robust evaluation experiments can serve as a comparative testbed for future analyses. 

\section{Datasets}
We use four fMRI datasets that have been collected by different researchers (see Table\,\ref{tbl_braindata}). All datasets use English language stimuli and the participants are native speakers. Standard fMRI preprocessing methods such as motion correction, slice timing correction and co-registration to an MNI template had already been applied.  

\subsection{Isolated stimuli}
We use two datasets that work with isolated stimuli. The stimuli are not related and can be presented in varying order to the participants. Each stimulus is represented with only a single brain activation vector by averaging over several scans obtained during the presentation of the stimulus.  

\paragraph{Words} For the \textsc{Words} dataset, 9 participants were shown a word paired with a line drawing of the object denoted by the word and were instructed to think about the properties of the object \cite{Mitchell2008}. Six scans were taken during the presentation of each word. The scans were temporally detrended and smoothed. The activation values were normalized by computing the percent signal change relative to the fixation condition. Scans and stimuli were aligned with an offset of 4 seconds to account for the haemodynamic delay. The brain activation for each word is calculated by taking the mean over the six scans. 
\paragraph{Stories} For the \textsc{Stories} dataset, 30 participants were reading 40 short personal stories that had been collected from weblogs \cite{Dehghani2017}. The stories consisted of 11 sentences on average  and were presented in three consecutive batches on a screen. The dataset also contains data for Farsi and Chinese stories, but for the sake of comparison, we focus on the English subset here. The scans were preprocessed with detrending, temporal smoothing and spatial smoothing. The activation values were normalized by calculating z-scores with respect to the fixation condition. The authors then discretized the continuous story stimulus by calculating the mean over all story scans. We exclude subject 30 from the data because the voxel values are all zero. 

\subsection{Continuous stimuli}
Humans process language incrementally and in context. In order to simulate a more naturalistic language setting, recent approaches to brain encoding use continuous stimuli and analyze the fMRI scans as a sequence of responses.   

\paragraph{Harry} For the \textsc{Harry} dataset by \citet{Wehbe2014}, 8 participants read chapter 9 of \textit{Harry Potter and the Sorcerer's stone} \cite{Rowling}. The story was split into four blocks and presented word by word on a screen. Each word was displayed for 0.5 seconds and an fMRI scan was taken every 2 seconds. We follow their protocol and apply detrending and temporal smoothing, but do not smooth spatially because it did not have an effect on the results in pilot experiments. 

\paragraph{Alice} For the \textsc{Alice} dataset by \citet{Brennan2016}, 27 participants were listening to an audio recording of the first chapter of \textit{Alice in Wonderland} \cite{Caroll}. The published data contains the preprocessed signal averaged for 6 regions of interests defined using functional and anatomical criteria. The raw signal is not available.

\section{Encoding Model}
The fMRI data is obtained by measuring the so-called blood-oxygenation level dependent (BOLD) response. This signal indicates the level of oxygen in the blood (approximated by its magnetic susceptibility) and an increased BOLD response in an area of the brain is interpreted as increased neuronal activity in this region. In order to analyze the response, the brain is fragmented into stacked voxels which are cubes of constant size (e.g. 3x3x3 mm). The response thus consists of a three-dimensional matrix with activation values for each voxel. This matrix is flattened into a one-dimensional vector $\mathbf{v}$. In the brain encoding approach, the goal is to predict $\mathbf{v}$ given the stimulus $\mathbf{s}$ that was presented when measuring the response. 

\paragraph{Mapping model}
A multiple linear ridge regression model is usually applied as encoding model to learn the response pattern $\mathbf{v_n}{\in}\mathbb{R}^m$ for stimulus $\mathbf{s_n}{\in}\mathbb{R}^d$ on a training set $V{\in}\mathbb{R}^{m{\times}n}$ of responses to $n$ other stimuli.\footnote{Whether a linear model is a plausible choice is debatable, but it has been most commonly used in previous work.} It requires a strong computational representation of the stimulus that reflects the relations between stimuli. The predictive power of this mapping model is evaluated on a set of held-out stimuli $S{\in}\mathbb{R}^{d{\times}n}$. The mapping model learns a separate regression equation for every voxel $v_i$ which is fitted by learning a weight $w_{d}$ for each dimension $\mathbf{s_d}$ of the stimulus representations and the weights are regularized by the L2 norm. The cost function $f$ for learning the weight vector $\mathbf{w}$ for a voxel vector $\mathbf{v_i}$ is: 

\[\displaystyle f(\mathbf{v_i}) = \sum_{n=1}^N ({v_{i_n} -\sum_{d=1}^D{w_d \cdot s_{d_n}}})^2 + \lambda \sum_{d=1}^D{w_d}^2\]

\subsection{Language model}
The linguistic stimuli are represented using vectors obtained from a language model. 
Previous work has compared the performance of different language models for brain encoding tasks showing that contextual models like long short-term memory networks perform better than standard word-based representations \cite{Jain2018}. For a more robust comparison, we keep the language model constant for all datasets. We choose the \textit{Elmo} language model because it produces contextualized representations on the sentence level and performs very well on semantic tasks \cite{Peters2018}. \textit{Elmo} is based on a bi-directional long short-term memory network and it uses character-based representations of the input which makes it perform very well on out-of-vocabulary words. This is an important property for modeling fictional texts. We use a pre-trained pytorch version of \textit{Elmo} available on github.\footnote{\url{https://github.com/allenai/allennlp/blob/master/tutorials/how_to/elmo.md}}

For \textsc{Words}, we use the representations from the token layer. For all other datasets, we obtain contextualized representations from the first layer. We restrict the representation to the forward language model to simulate incremental processing and obtain a 512-dimensional vector. We take the representation of the last token of each sentence and average over all sentences for each story in \textsc{Stories}. For the continuous stimuli, we feed the language model the whole chapter and extract the representation of the last token of the sequence which had been presented between the previous and the current scan.

\paragraph{Haemodynamic delay}
The fMRI signal measures a brain response to a stimulus with a delay of up to ten seconds \cite{miezin2000}. This delay needs to be considered when aligning stimuli with responses. Similarly to \citet{Mitchell2008}, we align scans to stimuli with a fixed offset of 4 seconds. The haemodynamic response decays slowly over a duration of several seconds. For continuous stimuli, this means that the response to previous stimuli will have an influence on the current signal. \citet{Wehbe2014} use a feature-based representation and learn different weights for stimuli occurring at previous time steps. In this approach, the number of features increases linearly with the number of time steps considered. In contextual language models, a representation is build up incrementally using recurrent connections. The representation of a word thus implicitly contains information from the previous context. As \textit{Elmo} processes language sentence by sentence, our context window comprises the current sentence up to the current word, but the number of dimensions remains constant. 

\subsection{Voxel selection}
The number of voxels in a brain varies with respect to the voxel size and the shape of the subject's brain. In the datasets used here, the number of voxels ranges from 20,000 to more than 40,000. The activity measured in many of these voxels is most likely not related to language processing, but might change due to physical processes like the noise perception in the scanner. In these cases, learning a mapping model from the stimulus representation to the voxel activation will not succeed because the stimulus has no influence on the variance of the voxel signal. Whole-brain evaluations of mapping models thus only have limited informative value. In previous work, different voxel selection models have been applied to analyze only a subset of interesting voxels.   
\citet{Wehbe2014} and \citet{Brennan2016} reduced the voxels by using previous knowledge about regions of interests.
Restricting the brain response to voxels that fall within a pre-selected set of regions of interests can be considered as a theory-driven analysis. 

\paragraph{Information-driven voxel selection} In contrast to the theory-driven region of interest analysis, \citet{KriegeskorteSearchLight} propose a more information-driven approach. So-called searchlight analyses move a sphere through the brain to select voxels (comparable to sliding a context window over text) and analyze the predictive power of the voxel signal within the sphere. \citet{Dehghani2017} and \citet{Wehbe2014} use this searchlight approach for the decoding task. In brain encoding, the predictive direction is reversed. The ability to predict voxel activation based on the stimulus is carefully interpreted as an indicator that processing the stimulus influences the activity in this particular voxel. For \textsc{Words}, \citet{Mitchell2008} analyze all six brain responses for the same stimulus and select 500 voxels that exhibit a consistent variation in activity across all stimuli.  \citet{Jain2018} calculate the model performance for a single voxel as the Pearson correlation between real and predicted responses on the test set and analyze voxels with a correlation above a threshold. \citet{Gauthier2018} recommend to evaluate voxels based on explained variance. We select the 500 most predictive voxels on the training set for \textsc{Words} by four selection methods: stability, Pearson correlation, explained variance, and random. 

\begin{table}[h]
\centering
\caption{The effect of voxel selection on the pairwise accuracy on \textsc{Words}. Accuracy and stable voxels are calculated as described in \cite{Mitchell2008}.  }
\label{tbl_Mitchell_vs}
\begin{tabular}{lccccc} 
\toprule
  Metric    &None       &Stable     &by $EV$          &by $R$           & Random \\
\midrule
  Cosine                &.57        & .65          &.67            &.56        & .57 \\
  Euclidean             &.57        & .66          &.67            &.56        & .57 \\
  Pearson               &.58        & .67          &.68            &.57        & .58\\
\bottomrule
\end{tabular}
\end{table}

\paragraph{Results of voxel selection} Table\,\ref{tbl_Mitchell_vs} shows the results for different voxel selection methods. It can be seen that voxel selection by explained variance performs on par with the selection of stable voxels. We had speculated that simply reducing the number of voxels might already lead to improvements because similarity measures tend to perform better in lower-dimensional spaces \cite{Aggarwal1998}, but a random selection of voxels has no effect. For the remainder of the paper, we report results on the 500 voxels that obtained the highest explained variance results on the training set unless indicated otherwise because the option of selecting stable voxels is not available for the other datasets.

\section{Evaluation experiments}
The voxel selection results show that a small experimental parameter can have a strong effect. We thus perform three experiments using different evaluation procedures: pairwise accuracy, voxel-wise evaluation, and representational similarity analysis. We repeat each experiment with a language model that assigns a random (but fixed) vector to each word to analyze the sensitivity of the evaluation metric. Random story representations are obtained by averaging over words. 

\subsection{Pairwise evaluation}
As the fMRI datasets are very small for machine learning purposes, \citet{Mitchell2008} introduced an evaluation procedure that maximizes the training data. Given a set of $n$ samples, a mapping model is trained on $n{-}2$ samples and tested on the two remaining samples. \citet{Mitchell2008} call this procedure leave-two-out cross-validation, but it differs from standard cross-validation setups because each sample occurs $n$ times in the test set leading to $\binom{n}{2}$ different models. The performance is evaluated by calculating the pairwise accuracy over all models. 

A pair of two test samples $(s_1,s_2)$ is considered to be classified correctly if the model prediction $p_1$ is more similar to the true target $s_1$ than to $s_2$, and $p_2$ is more similar to $s_2$. This general idea of pairwise accuracy has been implemented in different ways. The applied similarity metrics $f$ are cosine similarity \cite{Mitchell2008}, euclidean similarity \cite{Wehbe2014}, and Pearson correlation \cite{Bulat2017,Pereira2018}.
The prediction for a pair can be considered to be correct by comparing the summed similarity of the correct alignments with the false alignments \cite{Mitchell2008,Dehghani2017,Bulat2017}. \citet{Wehbe2014} and \citet{Wehbe2014align} calculate the accuracy by comparing the predictions only for the first sample. A stricter interpretation of the pairwise accuracy would only consider the prediction to be correct, if both samples are correctly matched to their prediction. We refer to the different interpretations as \textit{sum match} (1), \textit{single match} (2), and \textit{strict match} (3):

\begin{align}
&f(s_1,p_1) + f(s_2,p_2) > f(s_1,p_2) + f(s_2,p_1) \\
&f(s_1,p_1) > f(s_1, p_2)\\
&f(s_1,p_1) > f(s_1, p_2) \land f(s_2,p_2) > f(s_2, p_1)
\end{align}

\paragraph{Experimental setup}
 We calculate the pairwise accuracy for all four datasets, for the two similarity metrics cosine and euclidean and for the three match definitions sum, single, and strict. The leave-two-out evaluation only works well for isolated stimuli as in \textsc{Words} and \textsc{Stories}. For the continuous stimuli, we perform standard cross-validation. The \textsc{Harry} data can be split into four folds according to the experimental blocks and for the \textsc{Alice} data we determined 6 folds. The predictions for each fold are then paired with a randomly selected sample. We set a distance constraint between the two samples of at least 20 timesteps to avoid overlapping response patterns. For each sample, we average the result over 1000 random pairs as in \citet{Wehbe2014}.

\begin{table}
\centering
\caption{Pairwise accuracy results measured with cosine similarity, Euclidean similarity, and Pearson correlation and different match definitions averaged over all subjects. The results for the random language model are indicated in brackets. }
\begin{tabular}{llrrrrrrrrr} 
\toprule
&&\multicolumn{4}{c}{Encoding Model (Random LM)}   \\
 & Match& \textsc{Words}  & \textsc{Stories}  &\textsc{Alice} & \textsc{Harry}     \\
\midrule
\multirow{3}{*}{Cosine}
& Sum           &.67    (.54)&.57    (.53)&.54    (.53)&.50    (.49)\\
& Single        &.60    (.53)&.53    (.53)&.53    (.51)&.49    (.49)\\
& Strict        &.26    (.13)&.14    (.02)&.28    (.27)&.25    (.24)\\
\midrule
\multirow{3}{*}{Euclidean}
& Sum           &.67    (.53)&.56    (.53)&.53    (.53)&.50    (.49)\\
& Single        &.59    (.50)&.51    (.50)&.52    (.51)&.50    (.49)\\
& Strict        &.24    (.08)&.11    (.02)&.17    (.11)&.12    (.07)\\
\midrule
\multirow{3}{*}{Pearson's R}
& Sum           &.68    (.53)&.56    (.54)&.53    (.53)&.50    (.50)\\
& Single        &.61    (.53)&.52    (.52)&.52    (.52)&.50    (.49)\\
& Strict        &.26    (.10)&.11    (.02)&.27    (.27)&.25    (.24)\\
\bottomrule
\end{tabular}
\label{tbl_pair_results}
\end{table}

\begin{figure}
\centering
  \includegraphics[width=0.75\textwidth]{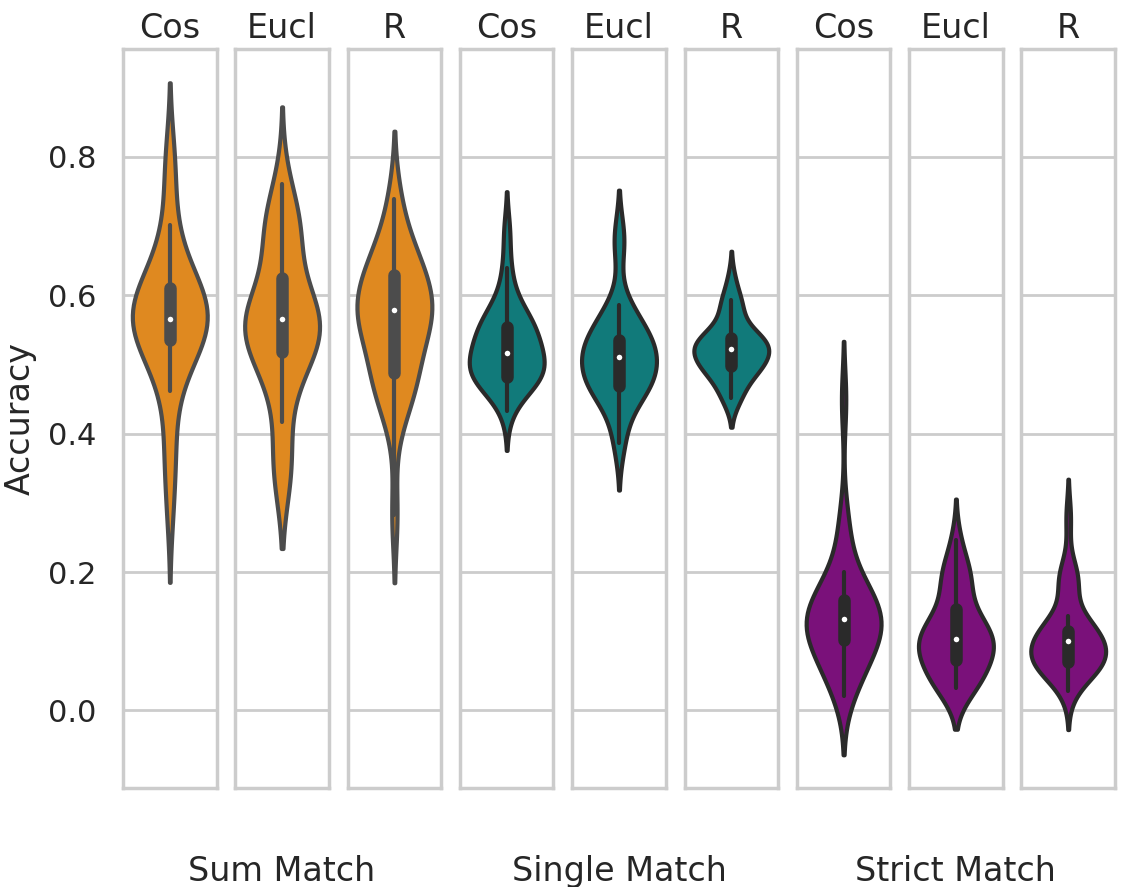}
\caption{Violin plot for the pairwise accuracy results for all subjects in \textsc{Stories} for each evaluation metric.}
  \label{fig:violinplot}
\end{figure}

\paragraph{Results}
The results in Table\,\ref{tbl_pair_results} are averaged over all subjects. It can be seen that the differences between the three similarity metrics and the sum and the single match are very small. The strict match is consistently more rigorous than the other match types. This indicates that both predictions would often be matched to the same stimulus when ignoring the pairwise exclusivity constraint. We conclude that the other two match types tend to slightly overestimate the discriminability of the stimulus. We also note that the difference to the random language model is more pronounced for the strict match for \textsc{Words} and \textsc{Stories}. For these two datasets, the results vary strongly across subjects. Subjects 1,3 and 4 in \textsc{Words} yield high accuracy results (0.87, 0.87, 0.76 for the cosine sum match) whereas the prediction for subject 6 is below chance level. We provide violin plots in Figure\,\ref{fig:violinplot} for a better impression of the variance across subjects in \textsc{Stories}. Although the results are worse than for \textsc{Words}, the accuracy is quite high for some subjects (0.80, 0.78, 0.7). The results obtained the isolated are comparable to those reported previously by \citet{Mitchell2008} and \citet{Dehghani2017}. For the continuous stimuli, the encoding model is not able to learn a robust signal. \citet{Wehbe2014} reported better results for the \textsc{Harry} data, but they performed the decoding task. \citet{Brennan2016} did not report encoding or decoding results, but focused on correlating the fMRI signal with computational models for surprisal.

\subsection{Voxel-wise evaluation}\label{s:voxelwise}
The pair-wise distance measures are an abstraction over all voxels. A model that mostly predicts constant values and only varies a few indicative voxels could perform well. As the mapping model independently predicts each voxel, we can take a closer look at the predictability of each voxel. 
This procedure accounts for the assumption that not every voxel in our brain will be influenced by the stimulus. In previous work, prediction results have often been reported only over significant voxels. 

\begin{table}
\centering
\caption{Voxel-wise results for cross-validation when taking the \textbf{average} over voxels. The results are averaged over all folds and all subjects. The results for the random language model are given in brackets. }
\label{tbl_voxelwise_mean}
\begin{tabular}{lllll} 
\toprule
&&\multicolumn{3}{c}{Average}\\
Voxels \phantom{extra space} &Dataset& \multicolumn{1}{c}{$EV$} &  \multicolumn{1}{c}{$R^2$}&\multicolumn{1}{c}{$r^2 simple$}\\
\midrule
\multirow{3}{*}{Whole brain} 
&Words      &-.21\,(-.09)           &-.41\,(-.35)       &.01\,(.01)\\ 
&Stories    &-.05\,(\phantom{-}.00) &-.26\,(-.20)       &.02\,(.01)\\
&Harry      &-.34\,(-.05)           &-.27\,(-.05)       &.00\,(.00) \\
\midrule
\multirow{3}{*}{Top 500 on train}&
Words       &-.14\,(-.08)           &-.33\,(-.26)       &.07\,(.11)\\
&Stories    &-.07\,(\phantom{-}.00) &-.27\,(-.19)       &.04\,(.02)\\
&Harry      &-.43\,(\phantom{-}.01) &-.44\,(-.07)       &.00\,(.00)\\
\midrule
\multirow{3}{*}{Top 500 on test} 
&Words      &\phantom{-}.42\,(\phantom{-}.21)   &\phantom{-}.34\,(\phantom{-}.05)   &.51\,(.37) \\
&Stories    &\phantom{-}.41\,(\phantom{-}.11)   &\phantom{-}.34\,(\phantom{-}.08)   &.68\,(.67) \\
&Harry      &-.12\,(\phantom{-}.01)             &-.12\,(\phantom{-}.01)             &.02\,(.02) \\
\bottomrule
\end{tabular}
\end{table}
{
\setlength{\tabcolsep}{2pt}
\begin{table}
\centering
\small
\caption{Voxel-wise results for cross-validation when taking the \textbf{sum} over voxels. The results are averaged over all folds and all subjects. The results for the random language model are given in brackets. }
\label{tbl_voxelwise_sum}
\begin{tabular}{llrrr} 
\toprule
&&\multicolumn{3}{c}{Sum}\\
Voxels&Data& \multicolumn{1}{c}{$EV$} &  \multicolumn{1}{c}{$R^2$}&\multicolumn{1}{c}{$r^2 simple$}\\
\midrule
\multirow{3}{*}{Whole} 
&Words   &\phantom{1}-4,303.86\,(-1,996.81)  &\phantom{5}-8,443.10\,(\phantom{4}-5,620.26)   &\phantom{4,}250.37(\phantom{2,}184.33)\\
&Stories &-10,232.30\,(\phantom{1,9}-47.56)  &-54,643.66\,(-42,297.88)                       &4,878.07\,(2,792.88)\\
&Harry&  -10,700.14\,(-1,451.43)             &-10,813.87\,(\phantom{4}-1,464.57)             &\phantom{4,32}-6.29\,(\phantom{2,7}-3.10)\\
\midrule
\multirow{3}{*}{500 train}
&Words      &\phantom{2}-68.82\,(-39.84)                &-164.67\,(-129.35)                             &33.34\,(\phantom{1}-0.38)\\
&Stories    &\phantom{-26}0.00\,(\phantom{3}-0.55)      &-134.31\,(\phantom{1}-96.88)                   &21.36\,(\phantom{1-}9.43)\\
&Harry      &-215.45\,(-37.28)                          &-218.14\,(\phantom{1}-37.63)                   &\phantom{12}0.11\,(\phantom{1}-0.09)\\
\midrule
\multirow{3}{*}{500 test} 
&Words      &209.98\,(104.76)           &253.98\,(\phantom{1-}25.66)                &171.56\,(187.20)\\
&Stories    &204.90\,(\phantom{1}56.30) &339.33\,(\phantom{1-}39.52)                &171.08\,(334.99)\\
&Harry      &-58.66\,(\phantom{10}7.40) &-59.70\,(\phantom{1-2}7.23)&\phantom{1}10.41\,(\phantom{18}9.86)\\
\bottomrule
\end{tabular}
\end{table}
}

\paragraph{Experimental setup}
The explained variance ($EV$) and the coefficient of determination ($R^2$) are the most common metrics for evaluating linear regression. They measure the proportion of the variance in the dependent variable that is predictable by the model. The two metrics are closely related, but explained variance also accounts for the mean error. We use the implementation of these scores in the python library \textit{scikit-learn} \cite{scikit-learn}. \citet{Jain2018} calculate a different $r^2$ value: they multiply the Pearson correlation between the predictions and the observed activations for voxel $v_i$ with the absolute correlation ($r^2(v_i) = r_{v_i}{\times}\vert r_{v_i} \vert$). We refer to this measure as $r^2 simple$. They report the sum over all voxels averaged over all subjects. We calculate the voxel-wise results for all three metrics and combine them by averaging or summing over all voxels. We compare the results for the whole brain with a selection of the 500 best-performing voxels on the training and on the testing set respectively. Selection on the test set is not recommended, but added to compare previous work. 

\begin{figure}
\centering
  \includegraphics[scale=0.5]{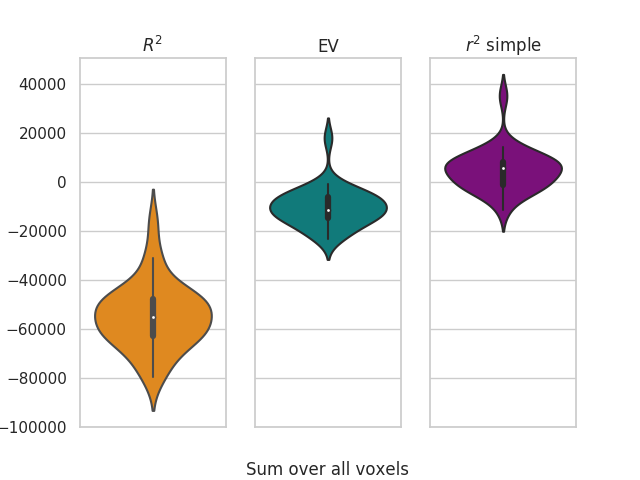}
\caption{Violin plots of the voxel-wise results (summed over all voxels) for all subjects in \textsc{Stories} for 500 voxels selected on the train and the test data.}
\label{fig:voxelwise_whole}
\end{figure}

\begin{figure}
  \centering
  \includegraphics[scale=0.65]{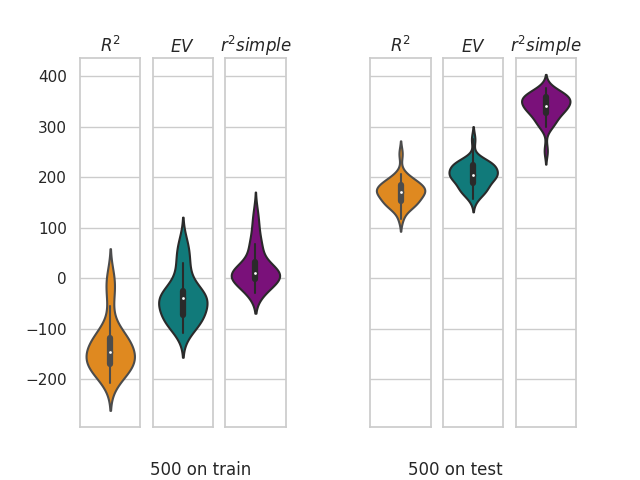}
\caption{Violin plots of the voxel-wise results (summed over all voxels) for all subjects in \textsc{Stories} for all voxels. Note the extreme change in the scale of the y-axis compared to Figure \ref{fig:voxelwise_whole} due to the number of voxels. }
\label{fig:voxelwise_top}
\end{figure}

\paragraph{Results}
Tables\,\ref{tbl_voxelwise_mean} and Tables\,\ref{tbl_voxelwise_sum} show the results for the voxel-wise evaluation averaged over all subjects . The metrics are presented averaged over all voxels in Table \ref{tbl_voxelwise_mean} and summed in Table \ref{tbl_voxelwise_mean}. It can be seen that the models are highly overfitted, so that we get much better results when voxels are directly selected on the test results than when they are pre-selected on the training data. In this setting, all metrics indicate that the encoding model is stronger than the random language model for \textsc{Words} and \textsc{Stories}. For the other settings, the explained variance and the $R^2$ are always negative. A value of zero for explained variance is obtained for a model that constantly predicts the mean. It is almost impossible to identify which one of two very negative models performs less bad based on this value alone. The averaged results vary less because we are averaging over all voxels, over all folds and over all subjects. Both, the inter-subject variance and the variance in voxel predictability are very high, so that positive and negative results cancel each other out. The $r^2 simple$ metric almost always returns a positive score. This might be a more satisfying result when evaluating the encoding quality; however, the metric also returns high positive scores for the random language model in some cases. Sum metrics depend on the number of voxels over which they are calculated.  For the whole brain analysis, averaged sum metrics are thus not interpretable in absolute terms because the number of voxels in the brain varies between subjects. For a better impression of the variance, we again provide violin plots for \textsc{Stories} in Figure\,\ref{fig:voxelwise_whole} for the whole brain analysis and in Figure \,\ref{fig:voxelwise_top} for selected voxels. We see that the results for the $r^2 simple$ metric are consistently better, but the extreme change on the x-axis indicates that sum scores should be interpreted with caution.

\begin{figure}[t]
\centering
\includegraphics[width=0.6\textwidth]{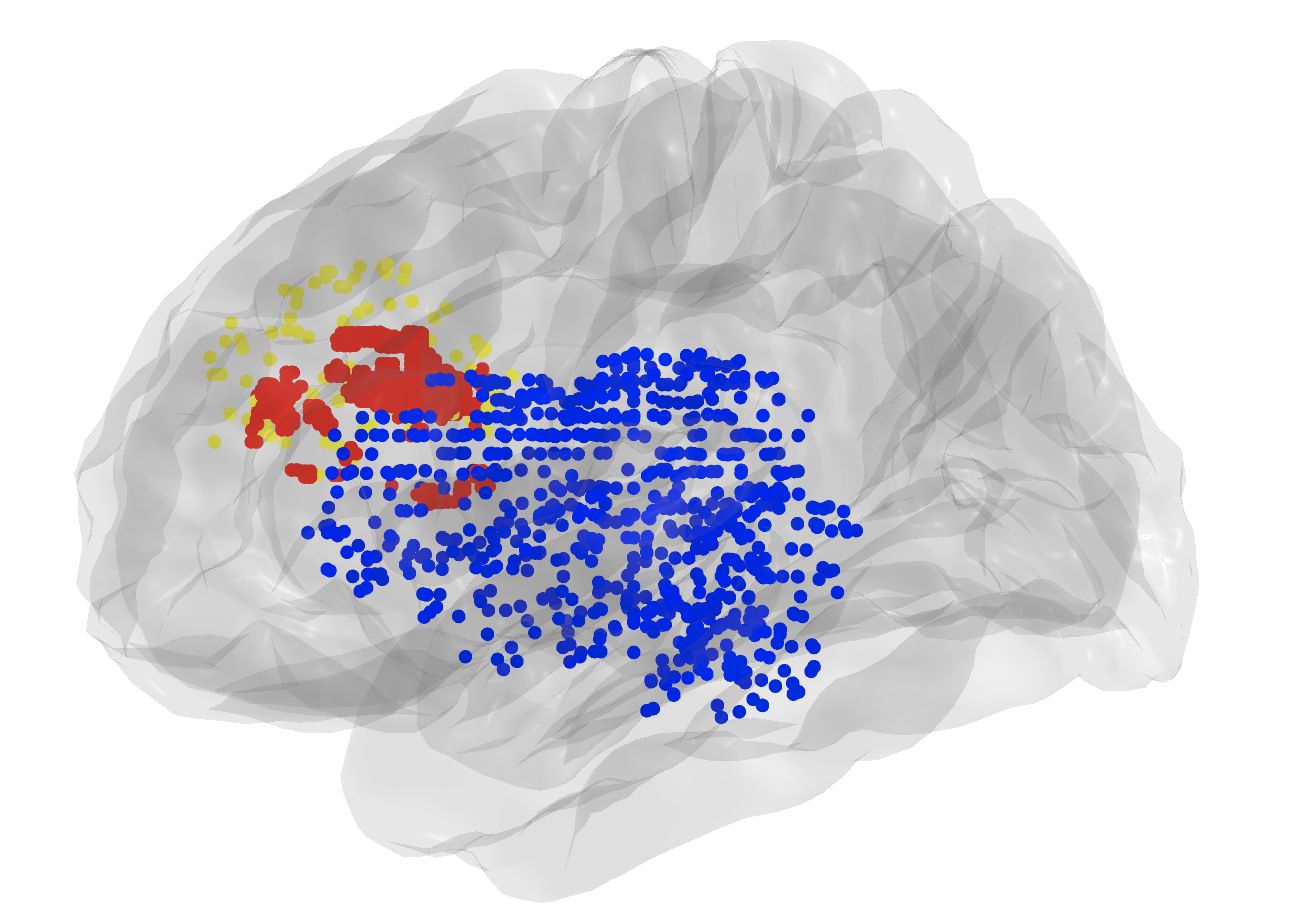}
 \caption{Predictive voxels for \textsc{Words} in blue, \textsc{Stories} in red and \textsc{Harry} in yellow.}
  \label{fig:voxel_plot} 
\end{figure}

\paragraph{Model-driven voxel selection}
We additionally determine the voxels with the highest explained variance on the test set when training on 80\% of the data. We set a threshold (0.3 for \textsc{Stories} and \textsc{Words}, 0 for \textsc{Alice}) and plot predictive voxels for the subjects for which we obtained highest accuracy in the pairwise comparison in Figure\,\ref{fig:voxel_plot}. The results are rather inconclusive. There is almost no overlap in the voxels and they are spread over several brain regions. This indicates that model-driven voxel information should only be interpreted on larger datasets.    


\subsection{Representational Similarity Analysis} 
The previous methods indicate that the continuous stimuli cannot be well encoded. In order to be able to attribute this flaw more directly to the language model, we perform representational similarity analysis \cite{Kriegeskorte2008} to compare the relations between brain activation vectors to the relations between stimulus representations without the intermediate mapping model.
The approach assumes that similar brain activation patterns are caused by strongly related stimuli. The quality of the computational representation of the stimuli can then be assessed by its ability to model these relations \cite{Anderson2016,Bulat2017,Xu2016}. As commonly performed in previous work, we measure the relations between vectors by the cosine distance and compare brain scans and representations by Spearman correlation and Pearson correlation. 

\paragraph{Results}
At first glance, the results in Table\,\ref{tbl_rsa_results} seem to confirm the impression that the encoding model performs better for the isolated stimuli. However, the same results can be obtained with the random language model. The random model can to a certain extent capture word identity (recall that the same random vector is assigned to different occurrences of the same word), so it might capture a relevant signal for the story stimuli, but this does not explain the results for the \textsc{Words} dataset with 60 different words. It can be seen that generally the more conservative rank-based Spearman correlation is much lower than the Pearson correlation. For the current setup, the representational similarity analysis results are unsatisfactory. However, the methodology largely reduces the number of parameters and facilitates the comparison of different computational models. We thus think that it could be a promising analysis method for future experiments. 
\begin{table}
 \caption{Results for representational similarity analysis calculated for the whole brain using Pearson correlation and Spearman correlation. The results for the random language model are indicated in brackets. }
 \label{tbl_rsa_results}
\begin{tabular}{lrrrr} 
\toprule
Metric &  \textsc{Words}  & \textsc{Stories}  &\textsc{Alice} & \textsc{Harry}     \\
\midrule

Spearman               &0.09 (0.05)   &0.08 (0.09)        &0.03   (0.01)   &0.00  (0.01)   \\
Pearson               &0.41 (0.44)    &0.19  (0.22)       &0.06  (0.02)   &0.06  (0.03)   \\
\bottomrule
\end{tabular}
\end{table}

\section{Discussion}
The setup of encoding experiments requires many modelling decisions for the stimulus representation, the stimulus--response alignment, the mapping model and its learning parameters, the noise reduction techniques for the brain responses, the voxel selection etc. Experimenting with a single dataset bears the danger of overfitting the experimental setup. We have seen that different evaluation metrics can interpret the predictive power of an encoding model very differently. Encoding results should thus always be compared to a reasonable baseline and hypotheses should be tested over several datasets. In this comparison, we intentionally restricted the experimental setup by choosing the same language model for all datasets. At this point, it remains unclear, whether the close to random results in many settings result from an unfortunate choice of the language model or from a noisy signal. Our experimental pipeline is modular and provides a useful testbed for future experiments with alternative stimuli representations. 

More sophisticated context models might increase the number of dimensions. From a machine learning perspective, most encoding experiments are problematic because the number of features is often higher than the number of samples. In addition, similarity metrics are known to sometimes behave unexpectedly when applied on high-dimensional data \cite{Aggarwal1998}. One could apply dimensionality reduction on the language representations, but these methods change the structure of the representation and make it difficult to derive cognitive insights for the original model. For future data collections, it would be important to obtain more data points from fewer subjects to facilitate more powerful pattern analyses. 

FMRI encoding is an intriguing, but also very challenging task because of the noisy signal. Within the current state of the art, even a tiny signal that is significantly different from chance, can be seen as a success. The pairwise estimation measures can present the results in a more pronounced way. However, as our analysis with the strict match have shown, the other match definitions tend to give an overly optimistic impression of the discriminability of the stimuli. A similar problem occurs, when summing the $r^2 simple$ value only over predictive voxels. We are convinced that in the long run, the field benefits from a more conservative estimate of the predictive power of the developed models.  

\section{Conclusions}
We have performed a robust comparison for language--brain encoding experiments and receive very diverse results for different evaluation metrics. It is our hope that our experimental framework can pave the way for future experiments to gradually determine the optimal encoding parameters. We plan to extend our experiments to the datasets by \citet{Pereira2018} and to other languages. We can already provide a set of practical recommendations for evaluation: 1.\,For the pairwise evaluation, it is helpful to additionally report the strict match to put the results in perspective. 2.\,Averaging over subjects is not very informative, violin plots can give a better impression of the variance. 3.\,For sum metrics, it is important to clearly specify the number of voxels that are taken into consideration. 4.\,Voxel selection methods should be transparently described and only performed on the training set because they have a strong effect on the results. 
\section*{Acknowledgements}
The work presented here was funded by the Netherlands Organisation for Scientific Research (NWO),
through a Gravitation Grant 024.001.006 to the Language in Interaction Consortium.
\renewcommand{\bibsection}{\section*{References}} 
\bibliographystyle{splncs}
\begingroup
  \ifluatex
  \else
    \microtypecontext{expansion=sloppy}
  \fi
  \small 
  \bibliography{brainlang}
\endgroup


\end{document}